\documentclass[conference]{IEEEtran}
\IEEEoverridecommandlockouts
\usepackage{booktabs}
\usepackage[table,xcdraw]{xcolor}
\usepackage{hyperref}
\usepackage{comment}
\usepackage{balance}
\usepackage{amsmath,amssymb,amsfonts}
\usepackage[utf8]{inputenc} 
\usepackage[T1]{fontenc}    
\usepackage{hyperref}       
\usepackage{url}            
\usepackage{booktabs}       
\usepackage{amsfonts}       
\usepackage{nicefrac}       
\usepackage{microtype}      
\usepackage{xcolor}         
\usepackage{graphicx}
\usepackage{subfig}
\usepackage{todonotes}
\usepackage{authblk}
\usepackage{tabularx}
\usepackage[inline, shortlabels]{enumitem}

\makeatletter
\def\input@path{{./sections/}}
\makeatother

\graphicspath{{figures/}}

\title{Advancing AI Challenges for the United States Department of the Air Force$^{*}$

\thanks{\textsuperscript{*}Invited Position Paper. Research was sponsored by the Department of the Air Force Artificial Intelligence Accelerator and was accomplished under Cooperative Agreement Number FA8750-19-2-1000. The views and conclusions contained in this document are those of the authors and should not be interpreted as representing the official policies, either expressed or implied, of the Department of the Air Force or the U.S. Government. The U.S. Government is authorized to reproduce and distribute reprints for Government purposes notwithstanding any copyright notation herein.

{\dag} Contributions to this work were made while the author was with MIT. A. Lancho is now with Universidad Carlos III de Madrid, Spain. G. Lee is now with Institute for Infocomm Research (I²R), A*STAR, Singapore. A. Weiss is now with Bar-Ilan University, Israel.
}}

\author{
Christian Prothmann,
Vijay Gadepally,
Jeremy Kepner,
Koley Borchard,
Luca Carlone, \\
Zachary Folcik, 
J. Daniel Griﬃth, 
Michael Houle, 
Jonathan P. How, 
Nathan Hughes, \\
Ifueko Igbinedion,
Hayden Jananthan, 
Tejas Jayashankar, 
Michael Jones,
Sertac Karaman, \\
Binoy G. Kurien, 
Alejandro Lancho\IEEEauthorrefmark{2},
Giovanni Lavezzi,
Gary C. F. Lee\IEEEauthorrefmark{2},
Charles E. Leiserson, \\
Richard Linares, 
Lindsey McEvoy, 
Peter Michaleas, 
Chasen Milner, 
Alex Pentland, 
Yury Polyanskiy, \\
Jovan Popovich,  
Jeffrey Price,
Tim W. Reid,
Stephanie Riley, 
Siddharth Samsi, 
Peter Saunders, \\
Olga Simek, 
Mark S. Veillette, 
Amir Weiss\IEEEauthorrefmark{2}, 
Gregory W. Wornell,
Daniela Rus,  
Scott T. Ruppel
\\
\\
Massachusetts Institute of Technology
}

\begin{document}

\maketitle

\begin{abstract}
The DAF-MIT AI Accelerator is a collaboration between the United States Department of the Air Force (DAF) and the Massachusetts Institute of Technology (MIT).
This program pioneers fundamental advances in artificial intelligence (AI) to expand the competitive advantage of the United States in the defense and civilian sectors. In recent years, AI Accelerator projects have developed and launched public challenge problems aimed at advancing AI research in priority areas. Hallmarks of AI Accelerator challenges include large, publicly available, and AI-ready datasets to stimulate open-source solutions and engage the wider academic and private sector AI ecosystem. This article supplements our previous publication, which introduced AI Accelerator challenges. We provide an update on how ongoing and new challenges have successfully contributed to AI research and applications of AI technologies.
\end{abstract}

\section{Introduction}

Established in 2019, the DAF-MIT AI Accelerator has successfully launched and funded projects that advance AI research in a broad range of areas to strengthen the technological lead of the DAF and to support society in general~\cite{Matheson.2019, Crichton.2019, DAFMITAIA}. The interdisciplinary project teams include researchers from MIT campus, MIT Lincoln Laboratory, and the DAF. In an effort to further the impact of AI Accelerator research and foster innovation, AI Accelerator projects have developed and rolled out various public challenge problems that engage the wider AI ecosystem. 
Public challenge problems—with well-defined problem sets, open and AI-ready datasets, and metrics—attract the ingenuity and expertise of academic and private sector organizations and individuals, amplifying innovation and strengthening ties within the AI community. In computer science, there are multiple examples of impactful public challenge problems, e.g., Graph Challenges~\cite{8547527}, MNIST~\cite{lecun1998mnist}, and ImageNet~\cite{deng2009imagenet}. AI Accelerator challenges are designed to advance AI research and also promote the transition of fundamental research towards the application of AI technologies.

\begin{figure*}[!ht]
  	\centering
    	\includegraphics[width=0.7\linewidth]{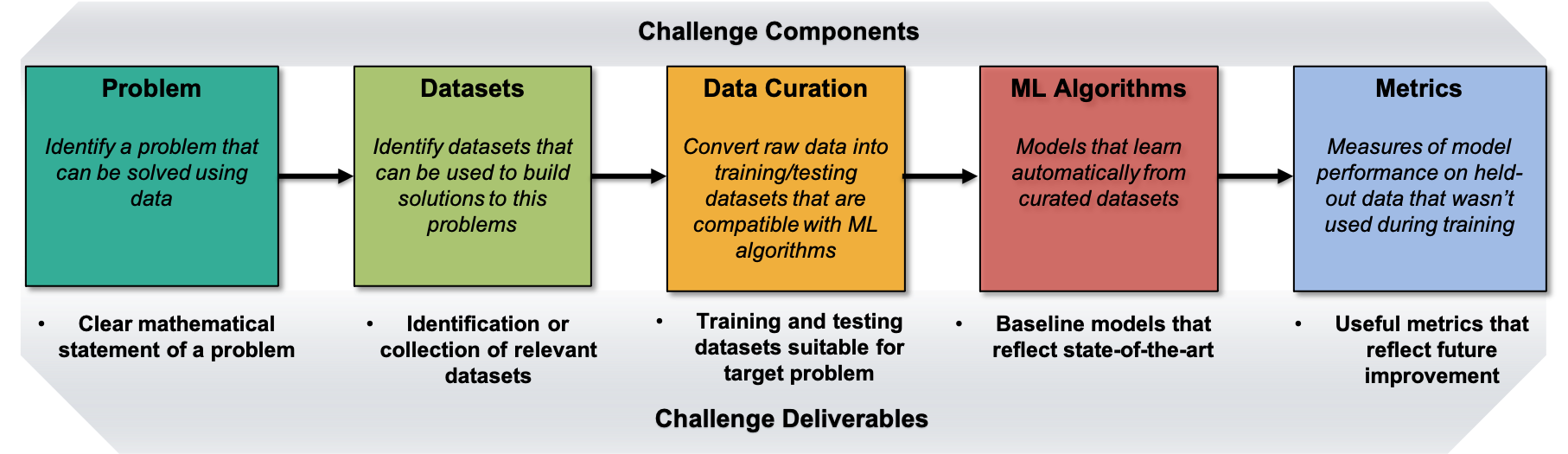}
	\caption{Components and description of challenges (reprinted from~\cite{9991948}).}
      	\label{fig:components}
\end{figure*}

We previously provided an overview~\cite{9991948} on the first set of AI Accelerator challenges, describing the approaches, datasets, and lessons learned, such as the importance of the development platform, data/code release, reproducible pipelines, and effectively engaging the user community. Successfully launching and running public challenge problems requires multiple other components as shown in Figure~\ref{fig:components}~\cite{9991948}.

In this article, we provide an update on AI Accelerator challenges and describe the establishment and progress of the following new and existing challenges:
\hfill
\begin{enumerate}[(1)]
	\item Data-Driven Radio-Frequency Signal Separation Challenge
	\item Tornado Network Challenge
	\item AI Innovation in Space Challenge
	\item Reinforcement Learning Challenge
    \item Anonymized Network Sensing Graph Challenge
    \item Datacenter Challenge
	\item Evaluating Long-Context LLM Architectures with Controlled Synthetic Benchmarks Challenge
\end{enumerate}
\hfill

\section{Data-Driven Radio-Frequency Signal Separation Challenge}
\label{sec:CPC}

\begin{figure}[!h] 
\centering
    \includegraphics[width=\columnwidth]{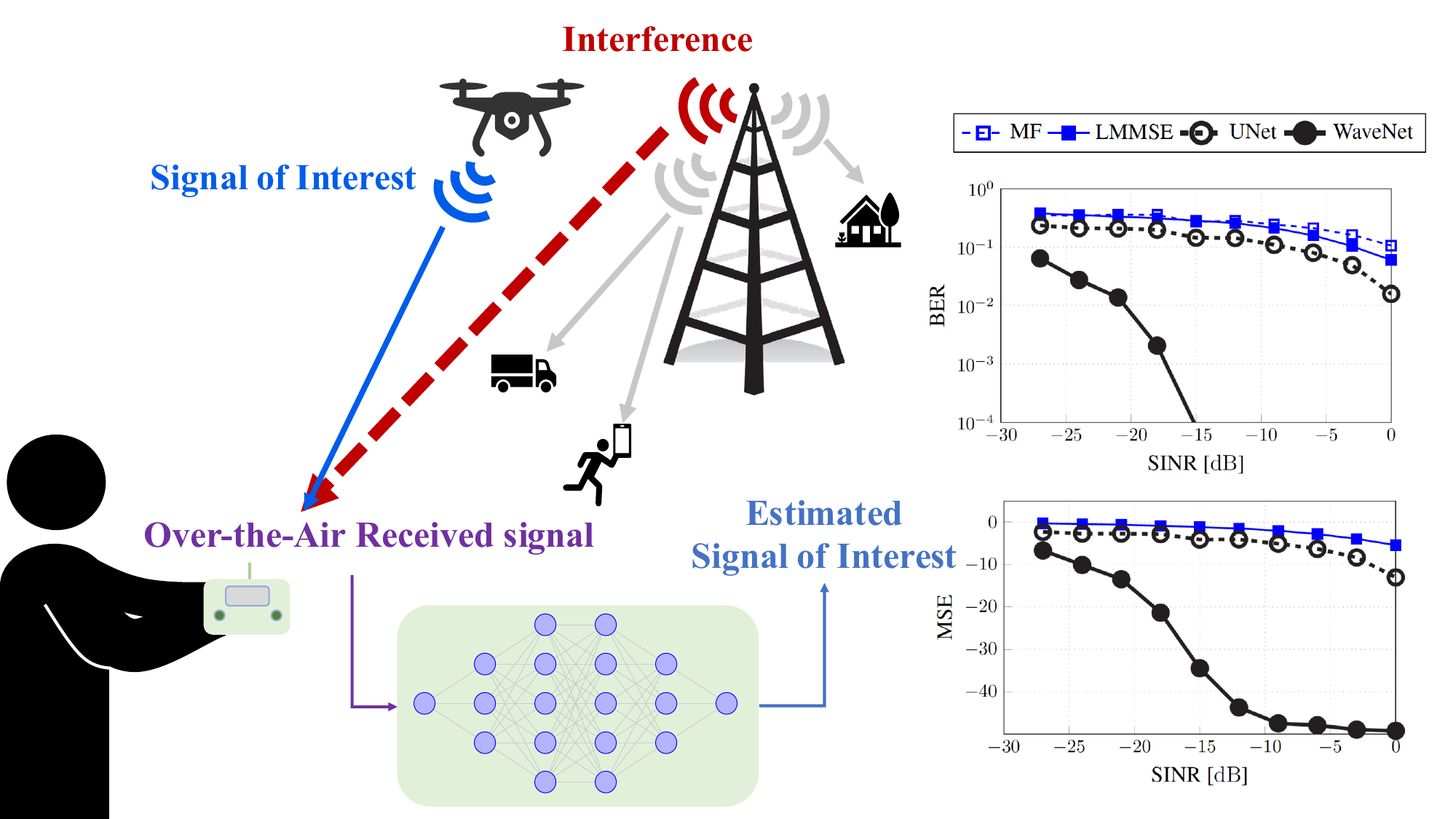} 
  \caption{Adapted from~\cite{Lancho25-04}. Left: Representative example of a co-channel interference scenario where multiple RF emitters operate in the same frequency band, leading to overlapping signals in time, frequency, and space. Right: Performance comparison for QPSK SOI in the presence of CommSignal5G1 interference. Both BER and MSE are plotted as a function of SINR for traditional baselines (MF, LMMSE) and our proposed deep learning-based methods (UNet, WaveNet).  
}
  \label{fig:ber_mse_commsignal5g1}
\end{figure}

\paragraph{Challenge Description}
As more radio-frequency (RF) systems operate in shared spectrum, unintentional interference is expected to become increasingly common. In many cases, such as the representative scenario depicted in the left side of Figure~\ref{fig:ber_mse_commsignal5g1}, the spatial resolution available at the receiver may be insufficient, and signals may overlap in time, frequency, and space. In such scenarios, effective methods must leverage the specific statistical structure of co-channel RF signals—waveforms that operate at the same time and frequency as the signal of interest (SOI)—through interference mitigation or signal separation approaches. The most common rejection tool, the matched filter (MF), treats all unwanted signals as if they were Gaussian noise and, while optimal for an additive white Gaussian noise channel~\cite{vantrees_01}, may fail when the interference itself is a non-Gaussian waveform overlapping the SOI. An alternative, when datasets are available, is to apply machine learning (ML) methods, which have already shown promise in the vision and audio domains. However, the RF domain poses unique challenges: signals are digitally synthesized, drawn from discrete alphabets, and exhibit both short- and long-term temporal statistical dependencies~\cite{Lancho25-04,lee2022exploiting,jayashankar2023score-neurips}. Moreover, the availability of labeled RF datasets has traditionally been limited, which has further slowed progress in this area. As a result, successful techniques from other domains often fall short~\cite{lee2023neural,datadrivenrf2024}. This challenge therefore benchmarks and drives the development of learning-based interference-rejection algorithms tailored to RF, aiming to recover the SOI at high fidelity and improve downstream tasks such as detection, demodulation, and decoding. To support this goal, we also released new RF signal datasets specifically designed to reflect the conditions and requirements of this problem.

\paragraph{Challenge Dataset}
The RF Challenge introduces a new dataset designed to reflect realistic interference scenarios where the signals overlap in time and frequency and the generative processes are often unknown. The dataset is publicly available at~\cite{rfchallenge} and includes several raw RF signal recordings, both over-the-air and generated in controlled lab environments. Most recordings are from the 2.4 GHz ISM band, with the exception of 5G waveforms generated via a cable setup and simulated channel models. 

In particular, four types of interference signals—EMISignal1, CommSignal2, CommSignal3, and CommSignal5G1—are provided only through recordings, with no access to their underlying generative processes. The first three were recorded over-the-air, while the 5G signal was captured in a wired setup with added impairments. Importantly, the ground truth signal structure is unknown even to the dataset creators, making these samples suitable for benchmarking interference rejection techniques under realistic assumptions.

\paragraph{Challenge Outcomes}
The right side of Figure~\ref{fig:ber_mse_commsignal5g1} compares the performance of our proposed deep learning-based interference rejection methods to classical baselines—MF and linear minimum mean squared error (LMMSE) estimation—for quadrature phase shift keying (QPSK) in the presence of CommSignal5G1 interference. The QPSK SOI is a single-carrier signal with root-raised cosine pulse shaping and high ($\times16$) oversampling.
For more implementation details, we refer the reader to~\cite[Sec. IV]{Lancho25-04}. Performance is measured as a function of the target signal-to-interference-and-noise ratio (SINR), using both bit error rate (BER) and mean-squared error (MSE) metrics.

Our deep-learning methods are based on two architectures tailored to RF signal properties, the UNet and the WaveNet~\cite[Sec. III.B]{Lancho25-04}. The UNet captures both local and long-range temporal structures through a multiscale architecture with downsampling and upsampling paths, connected via skip connections and equipped with comparatively long convolutional kernels. WaveNet, in contrast, preserves full temporal resolution using stacked residual blocks with dilated convolutions to capture long-term dependencies. As can be observed in the right side of Figure~\ref{fig:ber_mse_commsignal5g1}, both methods significantly outperform traditional baselines at low SINR levels. For instance, WaveNet achieves a BER near $10^{-3}$ at $-18\,\mathrm{dB}$ SINR, while LMMSE does not drop below $10^{-1}$. These results corroborate the potential of learning-based approaches for RF interference rejection, especially when signals overlap in time and frequency and traditional methods yield loose performance.

\section{Tornado Network Challenge}
\label{sec:tornet}
\paragraph{Challenge Description}

Tornadoes are among the most dangerous and least predictable weather phenomena, posing significant threats to life and property. Tornado threats have been on the increase, especially in the Southeast~\cite{Gensini.2018}, a region hosting several U.S. Department of Defense facilities~\cite{DoDInstallations.2024}. Accurate and timely detection of tornadic signatures in weather radar data is critical for issuing effective warnings and reducing false alarms. Traditional algorithms often rely on manually designed features and rule-based heuristics, which can struggle to generalize across storm types or handle ambiguous cases. In recent years, ML has emerged as a powerful tool for analyzing radar data directly. However, the development of robust models has been hampered by the lack of accessible, labeled datasets tailored to this problem. 

This challenge invites participants to develop ML models capable of identifying tornadoes from high-resolution radar data, leveraging the newly released Tornado Network (TorNet) dataset~\cite{veillette2025benchmark}. The primary motivation behind TorNet is to enable the broader research community to develop and benchmark ML algorithms for tornado detection. TorNet was constructed to address these issues by offering a curated collection of radar samples labeled according to whether a confirmed tornado was present.

\paragraph{Challenge Dataset}
The TorNet dataset comprises over 200,000 radar samples derived from 9 years of full-resolution, polarimetric WSR-88D (NEXRAD) level-II and level-III radar data. Each sample, called a "chip," includes multiple radar variables, such as reflectivity, radial velocity, and other variables captured across two elevation angles and four time steps spaced five minutes apart. TorNet retains the native polar format of the data, preserving spatial fidelity near the radar site. This level of detail enables the dataset to support a wide range of ML techniques, including deep learning models that can learn directly from raw radar imagery. The dataset includes both tornadic and non-tornadic cases, including borderline or ambiguous storms that resemble tornadoes in radar appearance. This design allows models trained on TorNet to confront realistic challenges such as class imbalance and false alarm reduction. Additionally, TorNet includes metadata for each sample, such as event IDs, storm severity, radar site, and EF ratings for confirmed tornadoes, enabling advanced tasks like data fusion and targeted analysis.

TorNet is designed for a broad array of potential use cases within meteorology and AI. It is particularly well-suited for classification tasks (e.g., distinguishing tornadic from non-tornadic radar signatures), time-series prediction (e.g., forecasting tornadogenesis), and interpretability research (e.g., explainable AI for radar imagery). Researchers can also use TorNet to test the robustness of their algorithms against rare events, handle uncertain or noisy labels, and benchmark new models against established baselines. The dataset includes deep learning baseline models and evaluation protocols, such as train/test splits that minimize data leakage, to foster reproducibility and fair comparisons. TorNet has already demonstrated its utility by enabling a convolutional neural network (CNN) to outperform traditional operational algorithms in detection accuracy. The dataset is publicly available and structured to be easily integrated into a variety of programming and ML frameworks, making it a valuable resource for both domain experts and newcomers working on severe weather prediction.

Instructions for downloading and working with TorNet can be found at~\cite{TorNetInstructions}.

\begin{figure}[!h]
\centering
  \includegraphics[width=\linewidth]{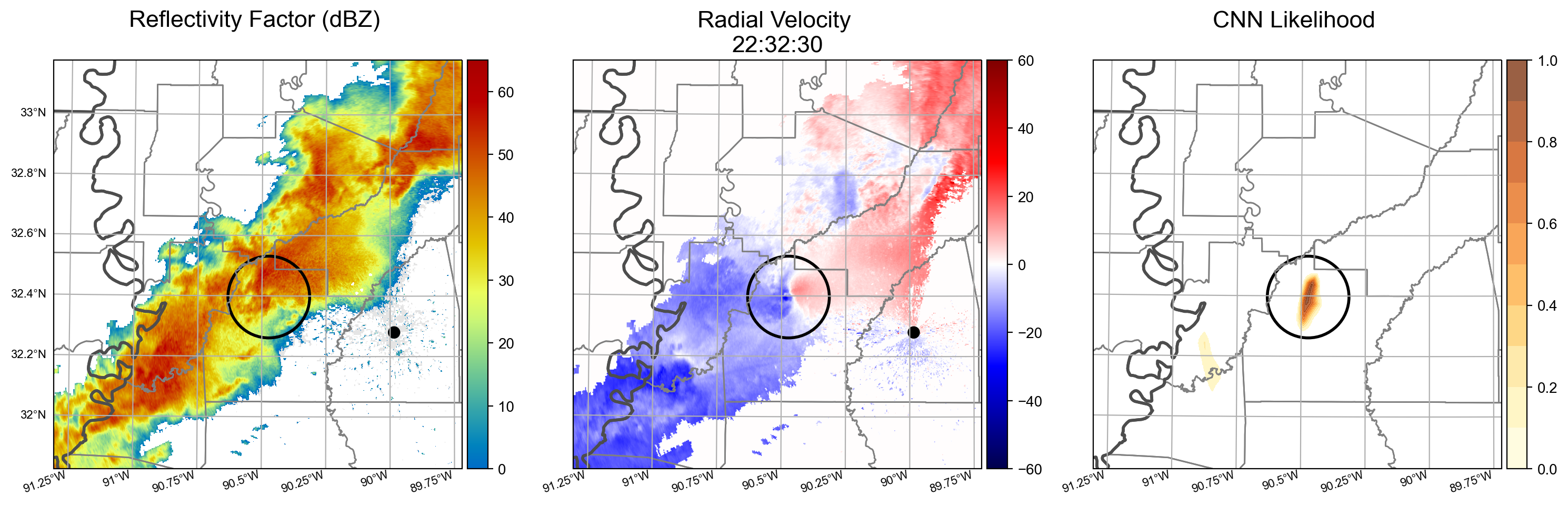}
  \caption{Reprinted from~\cite{veillette2025benchmark}: Tornado detection result for a storm impacting Dodge City Kansas, on 28 April 2024.  The left and middle panels show radar data of a line of storms that produced a tornado within the circled region.  A deep learning classifier trained using the TorNet dataset produced the likelihood image in the right panel, which highlights the location of the tornado.}
  \label{fig:sevir}
\end{figure}

\section{AI Innovation in Space Challenge: Satellite Pattern-of-Life Identification}
\label{sec:asdachallenge}

The 2024 MIT Astrodynamics, Space Robotics, and Controls Lab (ARCLab) Prize for AI Innovation in Space~\cite{siew2023ai,siew2024satellite,mit_challenge_2024_jas} was designed to stimulate the development and application of AI for characterizing satellite Pattern-of-Life (PoL) in Geostationary Earth Orbit (GEO). Participants were tasked with creating ML models capable of classifying satellite behavioral patterns and detecting key transition events (PoL nodes) from multivariate time-series astrometric data. The competition utilized a code-upload format on the EvalAI platform and involved a two-phase evaluation: Phase I focused on model performance (F2 score), and Phase II, for the top 10 teams, incorporated the quality of a technical write-up (assessing clarity, novelty, technical depth, reproducibility, and insights) into a composite score, with the F2 score still comprising 80\% of this final evaluation.

\paragraph{Challenge Dataset}
The primary dataset for the challenge was the Satellite Pattern-of-Life Identification Dataset (SPLID)~\cite{siew2023devkit}, which included 2,402 satellite trajectories, each spanning six months with a two-hour temporal resolution. These trajectories were sourced from both high-fidelity simulations (2,000 trajectories generated in collaboration with MIT Lincoln Laboratory to capture diverse behaviors, propulsion systems, and mission profiles) and historical data (402 trajectories, with 106 from Vector Covariance Messages (VCM) and 296 from Two-Line Elements (TLE)). Each simulated satellite was defined by unique physical parameters, initial orbital elements, and propulsion systems. The dataset provided astrometric parameters such as osculating orbital elements, geodetic positions, and Cartesian position and velocity vectors.

A critical component of SPLID was the expert-annotated, time-stamped PoL nodes, which marked transitions between different behavioral modes and specified the node type and propulsion system used in the subsequent mode. Behavioral modes included "not station-keeping" (NK), and station-keeping with "chemical" (CK), "electric" (EK), or "hybrid" (HK) propulsion. Node labels included "initiate drift" (ID), "adjust drift" (AD), and "initiate station-keeping" (IK), specified for both East-West (EW) and North-South (NS) directions. The dataset was divided into training, public test, and private test subsets, with the latter (comprising 502 state histories: 130 synthetic, 372 historical) used for final, unseen evaluation to prevent overfitting. The public dataset contained 1,900 distinct PoLs, representing 5,355 nodes and 8,957 behavioral modes.

\paragraph{Challenge Outcomes}
The 2024 AI Innovation in Space Challenge attracted significant international participation, with 126 registered teams, of which 35 actively submitted over 367 solutions. The top-performing teams demonstrated the effectiveness of diverse ML techniques (CNNs, long short-term memory networks (LSTMs), transformers, random forests, hybrid models), achieving F2 scores exceeding 0.9 on the partial test set and over 0.8 on the more challenging full private test set. Team Hawaii2024, for example, achieved an F2 score of 0.95 on the public test set and 0.81 on the private test set, ultimately ranking first overall with a composite score of 0.960. The results highlighted AI's ability to uncover subtle patterns in satellite behavior, but also exposed challenges, particularly in generalizing from predominantly simulated training data to real-world data (especially TLE-derived) with greater variability and noise, and in classifying complex behavioral modes like hybrid propulsion.

\begin{figure}[!h]
\centering
\includegraphics[width=\linewidth]{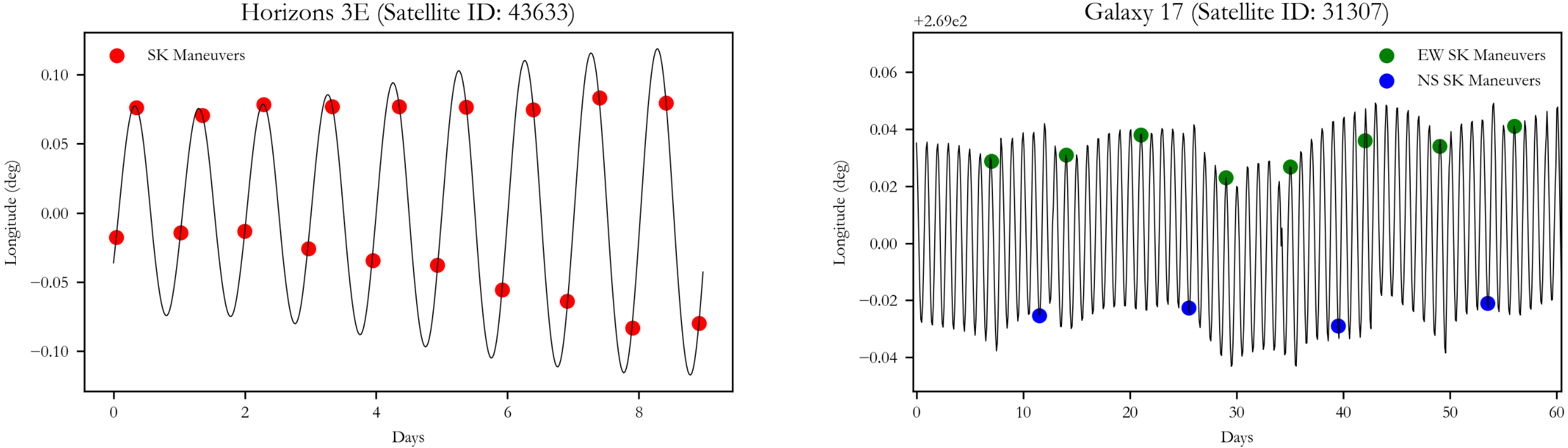}
\caption{An illustration of the classification problem central to the AI Challenge (reprint from~\cite{mit_challenge_2024_jas}): distinguishing different station-keeping (SK) strategies based on propulsion systems. 
The longitude history for \textit{Horizons 3E} demonstrates a composite SK routine, characterized by frequent, twice-daily maneuvers (red) typical of its electric propulsion system. 
In contrast, \textit{Galaxy 17} employs independent East-West (green) and North-South (blue) SK with its chemical propulsion system, resulting in far less frequent weekly and biweekly maneuvers. 
Participants' models were tasked with learning these distinct time-series patterns to classify behaviors and detect the underlying PoL nodes.}
\end{figure}

\section{Reinforcement Learning Challenge}
\label{sec:foodrace}
\paragraph{Challenge Description}

The Reinforcement Learning Challenge (“Race to Find the Food – First Come, First Served”) invited participants to develop autonomous reinforcement learning (RL) agents capable of searching for and identifying a target object in a virtual replica of MIT’s Stata Center. Framed around the light-hearted quest to find “pizza” before others did, the competition pushed participants to experiment with cutting-edge RL strategies, perception modules, and control systems in a dynamic 3D environment. With a simulated robot and limited time, participants raced to solve a high-impact vision and navigation problem that tested creativity and technical skill alike \cite{MIT-PSLAM}.

\begin{figure}[!h]
\centering
  \includegraphics[width=\linewidth]{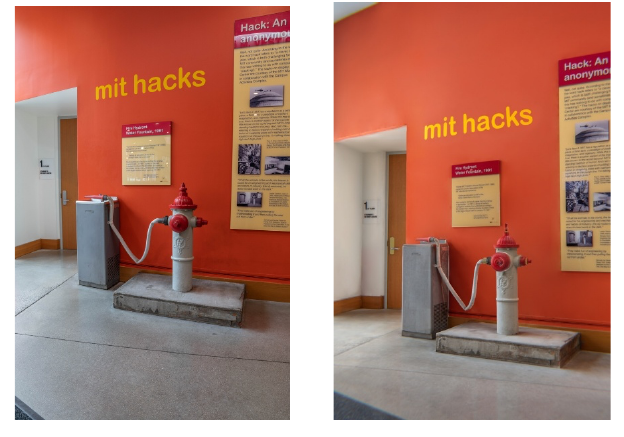}
  \caption{Example real image from Stata Center (left) and similar rendered image from FlightGoggles (right, \cite{FlightgogglesStata}.}
  \label{fig:foodrace}
\end{figure}

\paragraph{Challenge Dataset}
The environment was built using the FlightGoggles simulator, a high-fidelity platform originally designed for drone navigation~\cite{Flightgoggles}. Within this environment, the first floor of the MIT Stata Center was realistically rendered in 3D (Figure~\ref{fig:foodrace}). Each agent operated under Dubins vehicle dynamics and received three primary observations per step: a 256×192 grayscale camera image, a depth image, and its own pose (position and heading relative to the start). The agent's task was to locate a target placed randomly in the scene. Episodes ended when the agent found the target (within 3 meters and in its field of view), collided with a wall, or ran out of steps (capped at 1,000). To help participants get started, organizers provided a Gym-compatible interface, a baseline policy using Proximal Policy Optimization (PPO) from the RLlib library~\cite{RLlib}, and a ready-to-use simulation Docker environment. The baseline included a CNN-LSTM architecture that processed visual and pose data to output actions. Participants had full flexibility to adapt the reward model, learning algorithm, and training procedure using any preferred RL framework.
The challenge supported hands-on training by offering access to dedicated GPU-equipped workstations (Titan V) and maintained an open support channel via office hours, email, and GitHub issues.

\paragraph{Challenge Outcomes}
Three teams made up of students from MIT participated in the challenge in January 2024. The first-place team successfully identified 86 targets out of 100 test episodes—just one more than the second-place, highlighting the competitive nature of the event. Submissions were evaluated based on the number of targets located, with a tiebreaker based on the fewest average steps required to reach the target. Teams explored a variety of model strategies and policy architectures, showcasing not only deep technical expertise but also creative problem solving.

\section{Anonymized Network Sensing Graph Challenge}
\label{sec:betternet}

\paragraph{Challenge Description}

The costs of adversarial activity on networks are growing at an alarming rate. 90\% of Americans are now concerned about cyber attacks. The Department of the Air Force mission places it at the forefront of growing cyber challenges. Recent innovations in high-performance privacy-preserving network sensing and analysis offer new opportunities for obtaining the required observations to enable such architectures in the cyber domain. Using these network observations to pursue and counter adversarial activity requires the development of novel privacy-preserving AI-ready network sensors. 

The Anonymized Network Sensing Graph Challenge~\cite{jananthan2024anonymized} is part of the larger MIT/IEEE/Amazon Graph Challenge which encourages community approaches to developing new solutions for the analysis of graphs and sparse data derived from social media, sensor feeds, and scientific data to discover relationships between events as they unfold in the field. This particular challenge focuses on anonymized network traffic data at two levels, the first being PCAP (or "Packet Capture") files and the second being traffic matrices, seeking to enable large, open, community-based approaches to protecting networks~\cite{atkins2021improvised, atkins2021cooperation, demchak2021achieving, weed2022beyond, weed2023beyond}. At a high level, the challenge consists of processing PCAP files to extract and anonymize only the source and destination IP (Internet Protocol) addresses and then save those data to traffic matrices which are subsequently analyzed by calculating several specified network-theoretic statistics~\cite{soule2004identify, zhang2005estimating, mucha2010community, tune2013internet}. 
\begin{figure}[!h]
\centering
  \includegraphics[width=\linewidth]{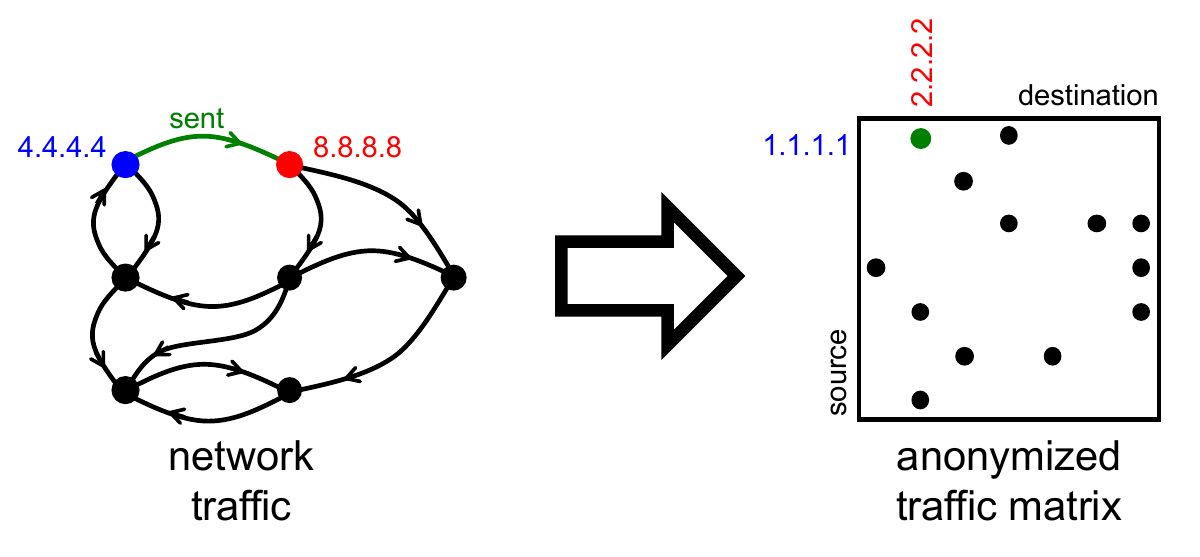}
  \caption{Graphical illustration of the conversion of network traffic into traffic matrices (reprint from~\cite{jananthan2024anonymized}).}
  \label{fig:NetworkToMatrix}
\end{figure}

Figure~\ref{fig:NetworkToMatrix} provides a high-level illustration of the conversion between network traffic data and traffic matrices. A collection of anonymized test data and reference implementations is provided for the challenge.

The complete process for performing the challenge consists of the following steps: 
\begin{enumerate}[(1)] 
	\item Read/stream each PCAP file.
	\item Extract source/destination IP addresses from each packet header. 
	\item Anonymize source/destination IP addresses in a consistent manner.
	\item Construct sequential traffic matrices. 
	\item Save the traffic matrices.
	\item Read and sum traffic matrices into a single large traffic matrix, and then perform analysis by calculating several network-theoretic statistics (see Figure~2 and Table~1 in~\cite{jananthan2024anonymized}).
\end{enumerate}

\paragraph{Challenge Dataset}

The anonymized test data available to challenge participants consists of two types, 
\begin{enumerate*}[(1)] 
	\item randomized and 
	\item anonymized real data collected from the Center of Applied Internet Data Analysis (CAIDA) darknet telescope~\cite{jananthan2024anonymized}.
\end{enumerate*} 
The CAIDA network traffic is collected and anonymized into traffic matrices in a process similar to that outlined for the challenge. 
Reference implementations in various programming languages are available to challenge participants. In the reference implementation, the source and destination IP addresses extracted from the PCAP files are anonymized using trusted sharing employing anonymization (e.g., CryptoPAN) that assumes the existence of an agreement prohibiting deanonymization. Additionally, traffic matrices are saved using SuiteSparse GraphBLAS matrix files, separated into $2^7$ {.tar} files each containing $2^6$ SuiteSparse GraphBLAS traffic matrices. Further information, including technical information about the datasets, the complete process for performing the challenge, and reference performance can be found in~\cite{jananthan2024anonymized}. 

\paragraph{Challenge Outcomes}
Several advances in both technologies and techniques have been motivated and supported by the Anonymous Network Sensing Graph Challenge. ~\cite{han2024extracting} combines the P4 network programming model and High-Level Synthesis (HLS) to achieve a processing rate of $\approx 95 \text{ Gbps}$ on field-programmable gate arrays.
\cite{samsi2025anonymized} uses the NVIDIA  RAPIDS ecosystem to achieve remarkable speedups for both graph construction and calculation of network theoretic statistics. ~\cite{lockton2025collaborative} looks at novel approaches to processing and analyzing network traffic data by shifting from external services to intrinsic capabilities of the operating system by combining DBOS~\cite{dbos2025} and network sensing.

\section{Datacenter Challenge}
\label{sec:dcc}
\paragraph{Challenge Description}
With the recent growth of data centers across the United States and the world, the MIT Data Center Challenge \cite{dcc} aims to promote research and development into growing issues surrounding the computing and infrastructure needs to support rapid AI growth~\cite{samsi2010matlab}. This growth, driven by the combination of increased parameter count of frontier AI models~\cite{zhao2023sustainable}, relative flattening of computing performance~\cite{reuther2023lincoln} and widespread adoption of AI~\cite{samsi2023words}, is well documented. However, research teams are often limited by the lack of public and well-documented datasets needed to look for potential optimization pathways. 

\paragraph{Challenge Dataset}

The primary dataset for the project~\cite{samsi2021supercloud} was made available via the Registry of Open Data supported by Amazon Web Services. The datasets consists of over two million detailed traces corresponding to computing workloads performed on the MIT SuperCloud computing system~\cite{reuther2018interactive} across the span of approximately 0.75 years. In addition to the computing traces, the dataset consists of filesystem metadata and building management system dataset (available with request to the dataset owners). Datasets are available via a public Amazon Web Services link with direct access to over 2TB of data. To further research into AI specific workloads, the data set also consists of labelled jobs and detailed GPU telemetry. Numerous articles such as~\cite{zhao2023sustainable, li2023clover, li2022ai, li2024sprout} also provide additional examples on usage and application of the dataset. Data anonymization was performed using techniques such as those described in~\cite{gadepally2015computing}.

\begin{figure}[!htpb]
    \centering
    \subfloat[GPU power draw]{\includegraphics[width=.3\linewidth]{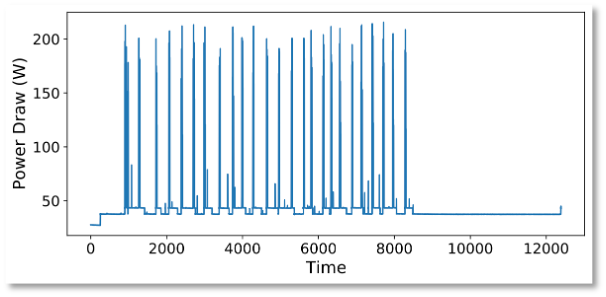}}
    \subfloat[GPU/CPU utilization]{\includegraphics[width=.4\linewidth]{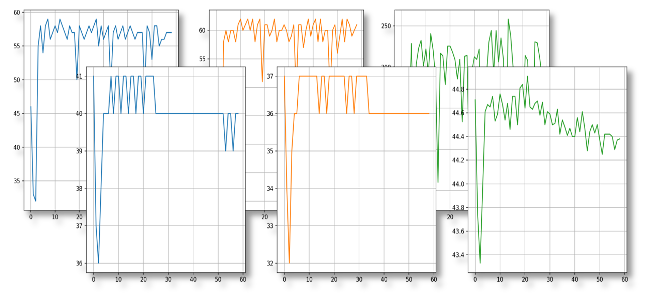}}
    \subfloat[GPU temperature]{\includegraphics[width=.3\linewidth]{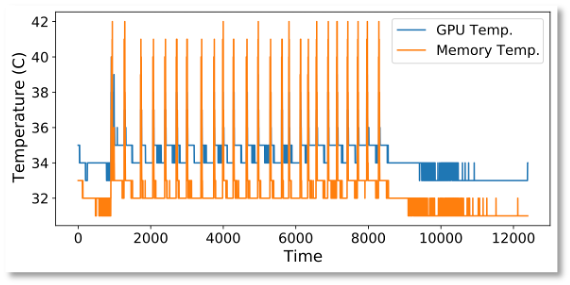}}
    \caption{Example time series data available as part of the MIT Supercloud dataset (reprint from~\cite{9991948}).}
    \label{fig:datacenter}
\end{figure}

\paragraph{Challenge Outcomes} 
The MIT Datacenter Challenge was officially released in conjunction with the IEEE International Parallel and Distributed Processing Society (IPDPS) conference. The associated workshop (\textit{AI for Data Center Optimization - ADOPT}) attracted dozens of participants across a variety of international institutions spanning academia, government and industry. To this date, the datacenter challenge remains one of the most detailed and relevant open-source workload datasets. Further, through a series of short-term courses at MIT, dozens of undergraduate students developed novel techniques for applications such as environmental impact reduction~\cite{li2023clover, li2023toward} and workload classification.

\section{Evaluating Long-Context LLM Architectures with Controlled Synthetic Benchmarks Challenge}
\label{sec:LLMArchitectures}

\paragraph{Challenge Description}
Current benchmarks for evaluating long-context capabilities of Large Language Models (LLMs) lack precise control over statistical properties of dependencies, making it difficult to systematically assess how different architectures handle varying types of long-range relationships~\cite{bai2025longbenchv2deeperunderstanding}. This challenge provides teams with a synthetic benchmark that can precisely replicate the dual scaling laws observed in natural human language—power-law decay in two-point mutual information between distant tokens and power-law growth in predictive information between text segments~\cite{chen2025l2mmutualinformationscaling}—while maintaining mathematically precise control over these information scaling properties. Teams will benchmark various architectures (Transformers, Mamba, hybrid models, etc.)~\cite{lieber2024jambahybridtransformermambalanguage, gu2023mamba} to determine which designs excel at different types of long-context tasks, identify fundamental limitations of current approaches, and explore how pretraining on language-realistic synthetic data might enhance models' long-context capabilities. Improving the long-range capabilities of LLMs can benefit the DAF in many applications such as complex question answering and event prediction.

\paragraph{Challenge Dataset}
The synthetic benchmark generates sequence data with controllable statistical properties that can precisely replicate the information scaling patterns observed in natural human language. The generator works by recursively composing layers of multivariate Gaussian distributions with structured correlations. By adjusting parameters like the reuse ratio between layers and local correlation patterns, teams can generate sequences that exhibit the exact same dual scaling laws found in literary corpora~\cite{chen2025l2mmutualinformationscaling} while maintaining full analytical control over the dependency structure.
Teams can create sequences with constant, logarithmic, power-law (sub-volume), and volume-law scaling of predictive information, including the specific sub-volume scaling that characterizes human language~\cite{chen2025l2mmutualinformationscaling}. The benchmark's key advantage is its ability to isolate and test specific aspects of long-range modeling that occur in real language. Unlike natural language benchmarks where multiple confounding factors are present~\cite{bai2025longbenchv2deeperunderstanding}, this controlled environment allows teams to determine whether architectural failures stem from inability to handle the fundamental information scaling laws of language versus other linguistic complexities.
The mathematical precision ensures reproducible results and enables principled comparison across different modeling approaches, while the language-realistic scaling provides direct insight into real-world performance expectations.

\begin{figure}[!h]
\centering
  \includegraphics[width=\linewidth]{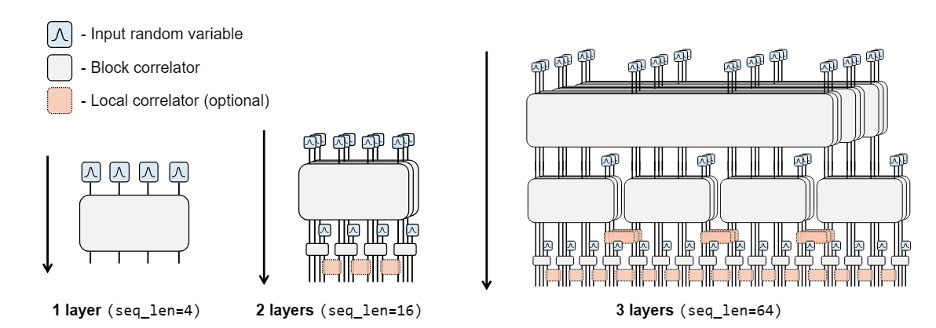}
  \caption{Pictorial illustration of the multi-scale structure of the synthetic dataset generator~\cite{L2Bench}.}
  \label{fig:LLMArchitectures}
\end{figure}

\paragraph{Challenge Outcomes}
Initial benchmarking across transformer variants, state-space models, and hybrid architectures reveals distinct performance profiles when tested on sequences with natural language scaling properties versus other dependency regimes~\cite{lieber2024jambahybridtransformermambalanguage, gu2023mamba}. Models that perform well on constant scaling can struggle with volume-law scaling requirements, suggesting fundamental architectural limitations. The controlled benchmark has enabled identification of specific points where different architectures begin to fail in modeling long-range dependencies, providing clear performance boundaries that correlate with real-world long-context language modeling capabilities~\cite{lieber2024jambahybridtransformermambalanguage}.
\section{Discussion}

MIT and the DAF established the DAF-MIT AI Accelerator in 2019 to synergize their joint expertise and resources~\cite{Matheson.2019},~\cite{Crichton.2019},~\cite{DAFMITAIA}. The AI Accelerator advances the field of AI to benefit the DAF, national security, and society at large by pioneering fundamental research and accelerating its transition into the DAF. By enlisting the wider expertise of academic and private organizations, the public challenges rolled out by the AI Accelerator support the AI Accelerator’s mission. Engaging the AI community through a series of challenges facilitates the formation of an ecosystem that drives the innovation and development of AI technologies and applications. Over the years, the AI Accelerator has created multiple challenges, instilling in the team a proficiency for setting up and running challenges, including developing reproducible processes for releasing datasets, sharing processes and datasets effectively across different challenges, and establishing “best practices”~\cite{Agreement.2020, Readiness.2023}. The AI Accelerator challenges described in this article show how harnessing the resourcefulness of the larger AI community can engender creative solutions to diverse hard problems.

\section*{Acknowledgments}

The authors wish to acknowledge the following individuals for their contributions and support: Daniel Andersen, LaToya Anderson, William Arcand, Sean Atkins, David Bader, David Baldsiefen, William Bergeron, David Bestor, Christopher Berardi, Bob Bond, Alex Bonn, Andrew Bowne, Stephen Buckley, Aydin Buluc, Chansup Byun, Zhuo Chen, K. Claffy, Cary Conrad, Timothy Davis, Chris Demchak, Phil Dykstra, Alan Edelman, Peter Fisher, Garry Floyd, Jeff Gottschalk, Daniel Grant, Dhruv Gupta, Oded Green, Isaac Haik, Thomas Hardjono, Chris Hill, David F. Jacobs, Daniel Jang, Kurtis Johnson, Tim Kraska, James Kurdzo, Miriam Leeser, Chris Long, Piotr Luszczek, Kirsten Malvey, Nathan Metzger, Lauren Milechin, Sanjeev Mohindra, Guillermo Morales, Julie Mullen, Dava Newman, Ritesh Patel, Roger Pearce, Heidi Perry, Ali Pinar, Sandeep Pisharody, Francois Porcher, Andrew Prout, Tony Reiss, Steve Rejto, Albert Reuther, Thomas G. Roberts, Victor Rodriguez-Fernandez, Antonio Rosa, Josh Rountree, Mark Sherman, Peng Mun Siew, Haley E. Solera, Binh Tran, Gabriel Wachman, Scott Weed, Michael Wright, Charles Yee, Christopher Yeung, and Marc Zissman.
\\
\\

\balance

\bibliographystyle{IEEEtran} 
\bibliography{references}

\begin{thebibliography}{10}
\providecommand{\url}[1]{#1}
\csname url@samestyle\endcsname
\providecommand{\newblock}{\relax}
\providecommand{\bibinfo}[2]{#2}
\providecommand{\BIBentrySTDinterwordspacing}{\spaceskip=0pt\relax}
\providecommand{\BIBentryALTinterwordstretchfactor}{4}
\providecommand{\BIBentryALTinterwordspacing}{\spaceskip=\fontdimen2\font plus
\BIBentryALTinterwordstretchfactor\fontdimen3\font minus
  \fontdimen4\font\relax}
\providecommand{\BIBforeignlanguage}[2]{{%
\expandafter\ifx\csname l@#1\endcsname\relax
\typeout{** WARNING: IEEEtran.bst: No hyphenation pattern has been}%
\typeout{** loaded for the language `#1'. Using the pattern for}%
\typeout{** the default language instead.}%
\else
\language=\csname l@#1\endcsname
\fi
#2}}
\providecommand{\BIBdecl}{\relax}
\BIBdecl

\bibitem{Matheson.2019}
\BIBentryALTinterwordspacing
R.~Matheson, ``M{IT} and {U.S. A}ir {F}orce sign agreement to launch {AI
  A}ccelerator,'' 2019. [Online]. Available:
  \url{https://news.mit.edu/2019/mit-and-us-air-force-sign-agreement-new-ai-accelerator-0520}
\BIBentrySTDinterwordspacing

\bibitem{Crichton.2019}
\BIBentryALTinterwordspacing
D.~Crichton, ``{MIT} and {A}ir {F}orce team up to launch {AI A}ccelerator,''
  2019. [Online]. Available:
  \url{https://techcrunch.com/2019/05/20/mit-and-u-s-air-force-team-up-to-launch-ai-accelerator}
\BIBentrySTDinterwordspacing

\bibitem{DAFMITAIA}
``{DAF-MIT AI A}ccelerator,'' [Online]. Available: \url{https://aia.mit.edu/},
  \url{https://www.aiaccelerator.af.mil}.

\bibitem{8547527}
S.~Samsi, V.~Gadepally, M.~Hurley, M.~Jones, E.~Kao, S.~Mohindra,
  P.~Monticciolo, A.~Reuther, S.~Smith, W.~Song, D.~Staheli, and J.~Kepner,
  ``Graph{C}hallenge.org: {R}aising the {B}ar on {G}raph {A}nalytic
  {P}erformance,'' in \emph{2018 IEEE High Performance extreme Computing
  Conference (HPEC)}, 2018, pp. 1--7.

\bibitem{lecun1998mnist}
\BIBentryALTinterwordspacing
Y.~LeCun, C.~Cortes, and C.~J.~C. Burges, ``The {MNIST} {D}atabase of
  {H}andwritten {D}igits,'' 1998. [Online]. Available:
  \url{http://yann.lecun.com/exdb/mnist/}
\BIBentrySTDinterwordspacing

\bibitem{deng2009imagenet}
J.~Deng, W.~Dong, R.~Socher, L.-J. Li, K.~Li, and L.~Fei-Fei, ``Image{N}et: A
  large-scale hierarchical image database,'' in \emph{2009 {IEEE} conference on
  computer vision and pattern recognition}.\hskip 1em plus 0.5em minus
  0.4em\relax {IEEE}, 2009, pp. 248--255.

\bibitem{9991948}
V.~Gadepally, G.~Angelides, A.~Barbu, A.~Bowne, L.~J. Brattain, T.~Broderick,
  A.~Cabrera, G.~Carl, R.~Carter, M.~Cha, E.~Cowen, J.~Cummings, B.~Freeman,
  J.~Glass, S.~Goldberg, M.~Hamilton, T.~Heldt, K.~W. Huang, P.~Isola, B.~Katz,
  J.~Koerner, Y.-C. Lin, D.~Mayo, K.~McAlpin, T.~Perron, J.~Piou, H.~M. Rao,
  H.~Reynolds, K.~Samuel, S.~Samsi, M.~Schmidt, L.~Shing, O.~Simek, B.~Swenson,
  V.~Sze, J.~Taylor, P.~Tylkin, M.~Veillette, M.~L. Weiss, A.~Wollaber,
  S.~Yuditskaya, and J.~Kepner, ``{D}eveloping a {S}eries of {AI Ch}allenges
  for the {U}nited {S}tates {D}epartment of the {A}ir {F}orce,'' in \emph{2022
  IEEE High Performance Extreme Computing Conference (HPEC)}, 2022, pp. 1--7.

\bibitem{vantrees_01}
H.~L.~V. Trees, \emph{Detection, Estimation, and Modulation Theory, Part
  I:}.\hskip 1em plus 0.5em minus 0.4em\relax New York, NY, USA: Wiley, 2001.

\bibitem{Lancho25-04}
A.~Lancho, A.~Weiss, G.~C.~F. Lee, T.~Jayashankar, B.~G. Kurien, Y.~Polyanskiy,
  and G.~W. Wornell, ``{RF} {C}hallenge: {T}he {D}ata-{D}riven {R}adio
  {F}requency {S}ignal {S}eparation {C}hallenge,'' \emph{IEEE Open J. Commun.
  Soc.}, vol.~6, pp. 4083--4100, Apr. 2025.

\bibitem{lee2022exploiting}
G.~C. Lee, A.~Weiss, A.~Lancho, J.~Tang, Y.~Bu, Y.~Polyanskiy, and G.~W.
  Wornell, ``Exploiting {T}emporal {S}tructures of {C}yclostationary {S}ignals
  for {D}ata-{D}riven {S}ingle-{C}hannel {S}ource {S}eparation,'' in \emph{IEEE
  Int. Workshop Mach. Learn. Signal Process. (MLSP)}, Aug. 2022.

\bibitem{jayashankar2023score-neurips}
T.~Jayashankar, G.~C. Lee, A.~Lancho, A.~Weiss, Y.~Polyanskiy, and G.~W.
  Wornell, ``Score-based {S}ource {S}eparation with {A}pplications to {D}igital
  {C}ommunication signals,'' in \emph{Advances Neural Inform.\ Proc.\ Syst.\
  (NeurIPS)}, New Orleans, LA, Dec. 2023.

\bibitem{lee2023neural}
G.~C. Lee, A.~Weiss, A.~Lancho, Y.~Polyanskiy, and G.~W. Wornell, ``On {N}eural
  {A}rchitectures for {D}eep {L}earning-{B}ased {S}ource {S}eparation of
  {C}o-{C}hannel {OFDM} {S}ignals,'' in \emph{IEEE Int. Conf. Acoust. Speech
  Signal Process.}, Jun. 2023.

\bibitem{datadrivenrf2024}
T.~Jayashankar, B.~Kurien, A.~Lancho, G.~C. Lee, Y.~Polyanskiy, A.~Weiss, and
  G.~Wornell, ``The {D}ata-{D}riven {R}adio {F}requency {S}ignal {S}eparation
  {C}hallenge,'' in \emph{Proc. IEEE Int. Conf. Acoust., Speech, Signal
  Process. (ICASSP)}, Apr. 2024.

\bibitem{rfchallenge}
\BIBentryALTinterwordspacing
``{RF C}hallenge at {MIT D}atasets.'' [Online]. Available:
  \url{https://rfchallenge.mit.edu/datasets/}
\BIBentrySTDinterwordspacing

\bibitem{Gensini.2018}
V.~A. Gensini and H.~E. Brooks, ``{Spatial Trends in United States Tornado
  Frequency},'' \emph{npj Clim. and Atmospheric Science}, vol.~1, no.~1, p.~38,
  2018.

\bibitem{DoDInstallations.2024}
\BIBentryALTinterwordspacing
{U.S. Department of Defense}, ``{Military {I}nstallations},'' 2024. [Online].
  Available: \url{https://installations.militaryonesource.mil/view-all}
\BIBentrySTDinterwordspacing

\bibitem{veillette2025benchmark}
M.~S. Veillette, J.~M. Kurdzo, P.~M. Stepanian, J.~Y. Cho, T.~Reis, S.~Samsi,
  J.~McDonald, and N.~Chisler, ``A {B}enchmark {D}ataset for {T}ornado
  {D}etection and {P}rediction using {F}ull-{R}esolution {P}olarimetric
  {W}eather {R}adar {D}ata,'' \emph{Artificial Intelligence for the Earth
  Systems}, 2025.

\bibitem{TorNetInstructions}
\BIBentryALTinterwordspacing
``{T}or{N}et.'' [Online]. Available: \url{https://github.com/mit-ll/tornet}
\BIBentrySTDinterwordspacing

\bibitem{siew2023ai}
P.~M. Siew, H.~E. Solera, T.~G. Roberts, D.~Jang, V.~Rodriguez-Fernandez, J.~P.
  How, and R.~Linares, ``{AI SSA} {C}hallenge {P}roblem: {S}atellite
  {P}attern-of-{L}ife {C}haracterization {D}ataset and {B}enchmark {S}uite,''
  in \emph{Proceedings of the Advanced Maui Optical and Space Surveillance
  (AMOS) Technologies Conference}, 2023, p.~5.

\bibitem{siew2024satellite}
P.~M. Siew, H.~E. Solera, G.~Lavezzi, T.~G. Roberts, D.~Jang, D.~Baldsiefen,
  B.~Tran, C.~Yeung, K.~Johnson, N.~Metzger, F.~Porcher, I.~Haik \emph{et~al.},
  ``Satellite {P}attern-of-{L}ife {I}dentification {C}hallenge: {C}ompetition
  {D}esign and {R}esults,'' in \emph{Advanced Maui Optical and Space
  Surveillance (AMOS) Technologies Conference}, 2024, p. 135.

\bibitem{mit_challenge_2024_jas}
P.~M. Siew, H.~E. Solera, G.~Lavezzi, T.~G. Roberts, D.~Jang, D.~Baldsiefen,
  B.~Tran, C.~Yeung, K.~Johnson, N.~Metzger, F.~Porcher, I.~Haik,
  V.~Rodriguez-Fernandez, Z.~Folcik, J.~Price, J.~P. How, and R.~Linares,
  ``{AI} {C}hallenge for {S}atellite {P}attern-of-{L}ife {I}dentification:
  {D}ataset, {D}esign and {R}esults,'' \emph{The Journal of the Astronautical
  Sciences}, 2025.

\bibitem{siew2023devkit}
\BIBentryALTinterwordspacing
P.~M. Siew, H.~E. Solera, T.~G. Roberts, D.~Jang, V.~Rodriguez-Fernandez, J.~P.
  How, and R.~Linares, ``{SPLID D}evelopment {K}it,''
  \url{https://splid-devkit.readthedocs.io/en/latest/README.html}, 2023,
  accessed: 2024-07-21. [Online]. Available:
  \url{https://splid-devkit.readthedocs.io/en/latest/README.html}
\BIBentrySTDinterwordspacing

\bibitem{MIT-PSLAM}
\BIBentryALTinterwordspacing
``{MIT-PSLAM}.'' [Online]. Available: \url{https://github.com/mit-pslam}
\BIBentrySTDinterwordspacing

\bibitem{FlightgogglesStata}
\BIBentryALTinterwordspacing
``{F}light{G}oggles {E}nvironments.'' [Online]. Available:
  \url{https://flightgoggles.mit.edu/virtual-environments/stata-center}
\BIBentrySTDinterwordspacing

\bibitem{Flightgoggles}
W.~Guerra, E.~Tal, V.~Murali, G.~Ryou, and S.~Karaman, ``Flight{G}oggles:
  {P}hotorealistic {S}ensor {S}imulation for {P}erception-driven {R}obotics
  using {P}hotogrammetry and {V}irtual {R}eality,'' \emph{2019 IEEE/RSJ
  International Conference on Intelligent Robots and Systems (IROS)}, 2019.

\bibitem{RLlib}
\BIBentryALTinterwordspacing
``{RL}lib: {I}ndustry-{G}rade, {S}calable {R}einforcement {L}earning.''
  [Online]. Available: \url{https://docs.ray.io/en/latest/rllib/index.html}
\BIBentrySTDinterwordspacing

\bibitem{jananthan2024anonymized}
H.~Jananthan, M.~Jones, W.~Arcand, D.~Bestor, W.~Bergeron, D.~Burrill,
  A.~Buluc, C.~Byun, T.~Davis, V.~Gadepally, D.~Grant, M.~Houle, M.~Hubbell,
  P.~Luszczek, P.~Michaleas, L.~Milechin, C.~Milner, G.~Morales, A.~Morris,
  J.~Mullen, R.~Patel, A.~Pentland, S.~Pisharody, A.~Prout, A.~Reuther,
  A.~Rosa, G.~Wachman, C.~Yee, and J.~Kepner, ``Anonymized {N}etwork {S}ensing
  {G}raph {C}hallenge,'' in \emph{2024 IEEE High Performance Extreme Computing
  Conference (HPEC)}, 2024, pp. 1--8.

\bibitem{atkins2021improvised}
S.~Atkins and C.~Lawson, ``An {I}mprovised {P}atchwork: {S}uccess and {F}ailure
  in {C}ybersecurity {P}olicy for {C}ritical {I}nfrastructure,'' \emph{Public
  Administration Review}, vol.~81, no.~5, pp. 847--861, 2021.

\bibitem{atkins2021cooperation}
------, ``Cooperation amidst competition: cybersecurity partnership in the {US}
  financial services sector,'' \emph{Journal of Cybersecurity}, vol.~7, no.~1,
  2021.

\bibitem{demchak2021achieving}
C.~Demchak, ``Achieving {S}ystemic {R}esilience in a {G}reat {S}ystems
  {C}onflict {E}ra,'' \emph{The Cyber Defense Review}, vol.~6, no.~2, pp.
  51--70, 2021.

\bibitem{weed2022beyond}
S.~Weed, ``Beyond {Z}ero {T}rust: {R}eclaiming {B}lue {C}yberspace,'' Master's
  thesis, United States Army War College, 2022.

\bibitem{weed2023beyond}
S.~Atkins and C.~Lawson, ``Beyond {Z}ero {T}rust: {R}eclaiming {B}lue
  {C}yberspace with {AI},'' \emph{Cyber Defense Review}, vol.~7, no.~1, 2023.

\bibitem{soule2004identify}
A.~Soule, A.~Nucci, R.~Cruz, E.~Leonardi, and N.~Taft, ``How to identify and
  estimate the largest traffic matrix elements in a dynamic environment,'' in
  \emph{ACM SIGMETRICS Performance Evaluation Review}, vol.~32.\hskip 1em plus
  0.5em minus 0.4em\relax ACM, 2004, pp. 73--84.

\bibitem{zhang2005estimating}
Y.~Zhang, M.~Roughan, C.~Lund, and D.~L. Donoho, ``Estimating point-to-point
  and point-to-multipoint traffic matrices: an information-theoretic
  approach,'' \emph{IEEE/ACM Transactions on Networking (TON)}, vol.~13, no.~5,
  pp. 947--960, 2005.

\bibitem{mucha2010community}
P.~J. Mucha, T.~Richardson, K.~Macon, M.~A. Porter, and J.-P. Onnela,
  ``Community {S}tructure in {T}ime-{D}ependent, {M}ultiscale, and {M}ultiplex
  {N}etworks,'' \emph{science}, vol. 328, no. 5980, pp. 876--878, 2010.

\bibitem{tune2013internet}
P.~Tune, M.~Roughan, H.~Haddadi, and O.~Bonaventure, ``Internet {T}raffic
  {M}atrices: {A} {P}rimer,'' \emph{Recent Advances in Networking}, vol.~1, pp.
  1--56, 2013.

\bibitem{han2024extracting}
Z.~Han, A.~Briasco-Stewart, M.~Zink, and M.~Leeser, ``Extracting {TCPIP}
  {H}eaders at {H}igh {S}peed for the {A}nonymized {N}etwork {T}raffic {G}raph
  {C}hallenge,'' in \emph{2024 IEEE High Performance Extreme Computing
  Conference (HPEC)}, 2024, pp. 1--6.

\bibitem{samsi2025anonymized}
S.~Samsi and V.~Gadepally, ``Anonymized {N}etwork {S}ensing {G}raph
  {C}hallenge,'' presented at NVIDIA GTC 2025.

\bibitem{lockton2025collaborative}
S.~Lockton, J.~Kepner, and H.~Jananthan, ``Collaborative awareness in {DBOS}:
  {A}n embedded approach to the {A}nonymized {N}etwork {S}ensing {G}raph
  {C}hallenge,'' submitted to IEEE HPEC 2025.

\bibitem{dbos2025}
\BIBentryALTinterwordspacing
``{DBOS} - {O}pen {S}ource {D}urable {E}xecution,'' https://www.dbos.dev/.
  [Online]. Available: \url{https://www.dbos.dev/}
\BIBentrySTDinterwordspacing

\bibitem{dcc}
\BIBentryALTinterwordspacing
``{D}ata {C}enter {C}hallenge.'' [Online]. Available:
  \url{https://dcc.mit.edu/}
\BIBentrySTDinterwordspacing

\bibitem{samsi2010matlab}
S.~Samsi, V.~Gadepally, and A.~Krishnamurthy, ``{MATLAB} for {S}ignal
  {P}rocessing on {M}ultiprocessors and {M}ulticores,'' \emph{IEEE Signal
  Processing Magazine}, vol.~27, no.~2, pp. 40--49, 2010.

\bibitem{zhao2023sustainable}
D.~Zhao, S.~Samsi, J.~McDonald, B.~Li, D.~Bestor, M.~Jones, D.~Tiwari, and
  V.~Gadepally, ``Sustainable {S}upercomputing for {AI}: {GPU} {P}ower
  {C}apping at {HPC} {S}cale,'' in \emph{Proceedings of the 2023 ACM Symposium
  on Cloud Computing}, 2023, pp. 588--596.

\bibitem{reuther2023lincoln}
A.~Reuther, P.~Michaleas, M.~Jones, V.~Gadepally, S.~Samsi, and J.~Kepner,
  ``Lincoln {AI} {C}omputing {S}urvey {(LAICS)} {U}pdate,'' in \emph{2023 IEEE
  High Performance Extreme Computing Conference (HPEC)}.\hskip 1em plus 0.5em
  minus 0.4em\relax IEEE, 2023, pp. 1--7.

\bibitem{samsi2023words}
S.~Samsi, D.~Zhao, J.~McDonald, B.~Li, A.~Michaleas, M.~Jones, W.~Bergeron,
  J.~Kepner, D.~Tiwari, and V.~Gadepally, ``From {W}ords to {W}atts:
  {B}enchmarking the {E}nergy {C}osts of {L}arge {L}anguage {M}odel
  {I}nference,'' in \emph{2023 IEEE High Performance Extreme Computing
  Conference (HPEC)}.\hskip 1em plus 0.5em minus 0.4em\relax IEEE, 2023, pp.
  1--9.

\bibitem{samsi2021supercloud}
S.~Samsi, M.~L. Weiss, D.~Bestor, B.~Li, M.~Jones, A.~Reuther, D.~Edelman,
  W.~Arcand, C.~Byun, J.~Holodnack \emph{et~al.}, ``The {MIT} {S}upercloud
  {D}ataset,'' in \emph{2021 IEEE High Performance Extreme Computing Conference
  (HPEC)}.\hskip 1em plus 0.5em minus 0.4em\relax IEEE, 2021, pp. 1--8.

\bibitem{reuther2018interactive}
A.~Reuther, J.~Kepner, C.~Byun, S.~Samsi, W.~Arcand, D.~Bestor, B.~Bergeron,
  V.~Gadepally, M.~Houle, M.~Hubbell, M.~Jones, A.~Klein, L.~Milechin,
  J.~Mullen, A.~Prout, A.~Rosa, C.~Yee, and P.~Michaleas, ``Interactive
  {S}upercomputing on 40,000 {C}ores for {M}achine {L}earning and {D}ata
  {A}nalysis,'' in \emph{2018 IEEE High Performance extreme Computing
  Conference (HPEC)}.\hskip 1em plus 0.5em minus 0.4em\relax IEEE, 2018, pp.
  1--6.

\bibitem{li2023clover}
B.~Li, S.~Samsi, V.~Gadepally, and D.~Tiwari, ``Clover: {T}oward {S}ustainable
  {AI} with {C}arbon-{A}ware {M}achine {L}earning {I}nference {S}ervice,'' in
  \emph{Proceedings of the International Conference for High Performance
  Computing, Networking, Storage and Analysis}, 2023, pp. 1--15.

\bibitem{li2022ai}
B.~Li, R.~Arora, S.~Samsi, T.~Patel, W.~Arcand, D.~Bestor, C.~Byun, R.~B. Roy,
  B.~Bergeron, J.~Holodnak \emph{et~al.}, ``{AI-E}nabling {W}orkloads on
  {L}arge-{S}cale {GPU-A}ccelerated {S}ystem: {C}haracterization,
  {O}pportunities, and {I}mplications,'' in \emph{2022 IEEE International
  Symposium on High-Performance Computer Architecture (HPCA)}.\hskip 1em plus
  0.5em minus 0.4em\relax IEEE, 2022, pp. 1224--1237.

\bibitem{li2024sprout}
B.~Li, Y.~Jiang, V.~Gadepally, and D.~Tiwari, ``Sprout: {G}reen {G}enerative
  {AI} with {C}arbon-{E}fficient {LLM} {I}nference,'' in \emph{Proceedings of
  the 2024 Conference on Empirical Methods in Natural Language Processing},
  2024, pp. 21\,799--21\,813.

\bibitem{gadepally2015computing}
V.~Gadepally, B.~Hancock, B.~Kaiser, J.~Kepner, P.~Michaleas, M.~Varia, and
  A.~Yerukhimovich, ``Computing on {M}asked {D}ata to improve the security of
  big data,'' in \emph{2015 IEEE International Symposium on Technologies for
  Homeland Security (HST)}.\hskip 1em plus 0.5em minus 0.4em\relax IEEE, 2015,
  pp. 1--6.

\bibitem{li2023toward}
B.~Li, R.~Basu~Roy, D.~Wang, S.~Samsi, V.~Gadepally, and D.~Tiwari, ``Toward
  {S}ustainable {HPC}: {C}arbon {F}ootprint {E}stimation and {E}nvironmental
  {I}mplications of {HPC} {S}ystems,'' in \emph{Proceedings of the
  international conference for high performance computing, networking, storage
  and analysis}, 2023, pp. 1--15.

\bibitem{bai2025longbenchv2deeperunderstanding}
\BIBentryALTinterwordspacing
Y.~Bai, S.~Tu, J.~Zhang, H.~Peng, X.~Wang, X.~Lv, S.~Cao, J.~Xu, L.~Hou,
  Y.~Dong, J.~Tang, and J.~Li, ``Long{B}ench v2: {T}owards {D}eeper
  {U}nderstanding and {R}easoning on {R}ealistic {L}ong-context {M}ultitasks,''
  2025. [Online]. Available: \url{https://arxiv.org/abs/2412.15204}
\BIBentrySTDinterwordspacing

\bibitem{chen2025l2mmutualinformationscaling}
\BIBentryALTinterwordspacing
Z.~Chen, O.~M. i~Comas, Z.~Jin, D.~Luo, and M.~Soljačić, ``L$^2${M}: {M}utual
  {I}nformation {S}caling {L}aw for {L}ong-{C}ontext {L}anguage {M}odeling,''
  2025. [Online]. Available: \url{https://arxiv.org/abs/2503.04725}
\BIBentrySTDinterwordspacing

\bibitem{lieber2024jambahybridtransformermambalanguage}
\BIBentryALTinterwordspacing
O.~Lieber, B.~Lenz, H.~Bata, G.~Cohen, J.~Osin, I.~Dalmedigos, E.~Safahi,
  S.~Meirom, Y.~Belinkov, S.~Shalev-Shwartz, O.~Abend, R.~Alon, T.~Asida,
  A.~Bergman, R.~Glozman, M.~Gokhman, A.~Manevich, N.~Ratner, N.~Rozen,
  E.~Shwartz, M.~Zusman, and Y.~Shoham, ``Jamba: {A} {H}ybrid
  {T}ransformer-{M}amba {L}anguage {M}odel,'' 2024. [Online]. Available:
  \url{https://arxiv.org/abs/2403.19887}
\BIBentrySTDinterwordspacing

\bibitem{gu2023mamba}
A.~Gu and T.~Dao, ``Mamba: {L}inear-{T}ime {S}equence {M}odeling with
  {S}elective {S}tate {S}paces,'' \emph{arXiv preprint arXiv:2312.00752}, 2023.

\bibitem{L2Bench}
Z.~Chen, O.~Mayne~i Comas, D.~Luo, and M.~Soljacic, ``L2{B}ench: {E}valuating
  {L}ong {C}ontext {L}ength {C}apabilities of {L}arge {L}anguage {M}odels with
  {P}hysics-{I}nspired {S}ynthetic {D}ata,'' \emph{in preparation}.

\bibitem{Agreement.2020}
D.~F. Jacobs, ``{D}ata {S}haring {A}greements: {T}emplate {A}ir {F}orce
  {S}ignable {A}greement and {T}emplate {A}ir {F}orce {C}lick-{T}hru
  {A}greement,'' [Online]. Available:
  \url{http://www.mit.edu/~kepner/AI-Accelerator/DataSharingAgreement-Signable.docx},
  http://www.mit.edu/~kepner/AI-Accelerator/DataSharingAgreement-ClickThru.docx,
  2020.

\bibitem{Readiness.2023}
\BIBentryALTinterwordspacing
V.~Gadepally and J.~Kepner, ``{S}imple {D}ata {A}rchitecture {B}est {P}ractices
  for {AI R}eadiness,'' 2023. [Online]. Available:
  \url{https://dspace.mit.edu/bitstream/handle/1721.1/152932/2023-11-03%20DataforAIReadiness.pdf?sequence=1&isAllowed=y}
\BIBentrySTDinterwordspacing

\end{thebibliography}

\end{document}